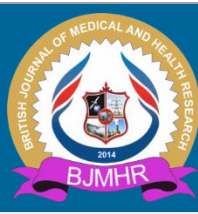

# Machine Learning Applications In Healthcare: The State Of Knowledge and Future Directions

**Mrinmoy Roy[1*], Sarwar J. Minar[2], Porarthi Dhar[3], A T M Omor Faruq[4]**
*1.Computer Science, Northern Illinois University, Dekalb, Illinois, USA.*
*2.Doctoral Student, Florida International University, Miami, Florida, USA.*
*3.BGC Trust Medical College, Chittagong, Bangladesh.*
*4. Analyst, Enterprise Project Management, CareFirst BCBS, Maryland, USA,*

## ABSTRACT

Detection of easily missed hidden patterns with fast processing power makes machine learning (ML) indispensable to today's healthcare system. Though many ML applications have already been discovered and many are still under investigation, only a few have been adopted by current healthcare systems. As a result, there exists an enormous opportunity in healthcare system for ML but distributed information, scarcity of properly arranged and easily explainable documentation in related sector are major impede which are making ML applications difficult to healthcare professionals. This study aimed to gather ML applications in different areas of healthcare concisely and more effectively so that necessary information can be accessed immediately with relevant references. We divided our study into five major groups: community level work, risk management/ preventive care, healthcare operation management, remote care, and early detection. Dividing these groups into subgroups, we provided relevant references with description in tabular form for quick access. Our objective is to inform people about ML applicability in healthcare industry, reduce the knowledge gap of clinicians about the ML applications and motivate healthcare professionals towards more machine learning based healthcare system.

**Keywords:** Machine Learning, Healthcare, Community Health, Telemedicine, AHC Screening

## INTRODUCTION

The application of Machine Learning is increasingly becoming increasingly common in various industries day by day. More strikingly, ML is shaping the face of almost every industry in today's world. ML can simply be defined as the capability of *machines* to learn from data by identifying patterns and to make predictions based on the knowledge gained in training stage. ML has become more popular as it can independently learn, adapt, grow, and develop with minimum human intervention [1].

The existing healthcare system is changing rapidly due to the fast-increasing number of patients and the complicated nature of their diseases. The most common problem in healthcare now is the shortage of staff in comparison with patient numbers but the quality of care depends on the quick response of nurses after the call button press. The introduction of modern technologies has also taken human expectations to an extremely high level. Moreover, diseases are changing their symptoms so much that they can easily be missed by the doctor's eye and become more malignant within a short period of time. So, to cope up with the ongoing challenges, healthcare systems need to reform their operations and management in such a way that staff and nurses can be allocated efficiently with proper resource management planning, devices can help healthcare professionals to make decisions (early disease detection), patients can get help without coming to hospital physically (telemedicine) and can be monitored continuously for any distress signal (fall monitoring, ICU monitoring). (Assisting human, increase human capacity)

Furthermore, the abundance of data in the form of text, audio, video, and the usage of modern EMR systems have made possible extensive data analysis. Nowadays availability of ICD-10 codes [2], patient's problem list and other biomedical data from laboratory [3] helps to create many short-terms and long-terms ML applications. Deep learning (DL), a subset of ML, simulates the human brain to produce neural network lies behind many modern healthcare services [4]. DL can process huge amounts of data within a noticeably short time and is very efficient in images, audio, and video data analysis. Moreover, Natural language processing (NLP), another branch of ML [5], enhances machine ability to understand text and spoken language and is intensively used in the health sector. NLP takes human output like writing or speaking as input and dictate machines to perform any required task. Another framework explainable AI (XAI) or explainable machine learning (XML) is becoming extremely popular due to making machine learning applications more understandable to end users. XAI has great importance in the healthcare industry because it characterizes ML model's accuracy, fairness, transparency and allows the users to trust ML models [6].

In this paper, we explored the existing innovative applications of ML in healthcare. After extensive research and interviewing some healthcare professionals in US, we found some important work of ML in healthcare, which we divided five (i.e., 5) main sections: ML in community level work, ML in risk management/ preventive care, ML in daily operational work, ML in remote care and ML in early detection. We further divide the main five categories into subcategories. First, Community level ML applications include Accountable Health Community (AHC) Screening and Disparity Index. Second, Risk Management/ Preventive Care ML applications include Hierarchical Condition Categories (HCC) Score Calculation, Mortality Analysis, and Surgical Risk Management. Third, Healthcare operation management ML applications include Resource Management, Patient Categorization, Patient Activity Monitoring, and Patience Experience Analysis. Fourth, Remote Care includes ML applications in Telemedicine. And fifth, Early Detection comprises of ML applications in Mental Health Detection, Early Disease Prediction, and Food Diet Planning. The overall data requirement for each group is shown in Figure 1. We also separated the important references of each subgroup and made group wise quick loop-up reference table for more exploration.

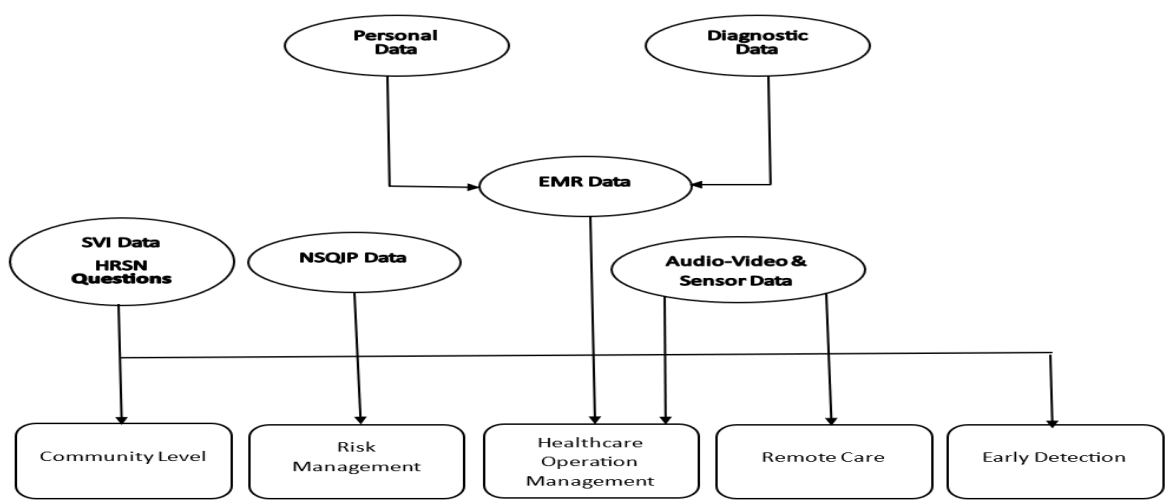

**Figure 1: Data Requirements for ML Applications on Healthcare**

## 2. Community level:

### 2.1. Accountable Health Community (AHC) Screening:

Health-related social needs (HRSNs) screening is playing a key role in many healthcare decisions making systems. HRSNs like hunger, homelessness, exposure to violence trigger different diseases and are considered as the major risk factors in health deterioration. Centers for Medicare & Medicaid Services (CMS) Center for Medicare and Medicaid Innovation (CMMI) made the Accountable Health Communities (AHC) model to inform patients' treatment plans to community services by screening Health-Related Social Needs (HRSN) [7]. Over seven million people will be screened using ten core domain questions in upcoming years by clinical delivery sites.

However, screening individuals with low-income and patients with higher risk of adverse health needs standard setting, particularly for healthcare. Moreover, primary care physicians and healthcare staff do not have enough confidence in individual screening process [8, 9] and depends on the patient's reporting. But HRSNs are strongly associated with social determinants of health (SDOHs) and electronic medical records (EMRs). Missed medical appointments, frequent emergency department (ED) visits, hospital readmissions [10, 11] helps to plan social intervention which in turn improves management of chronic conditions, increase access to preventive healthcare and reduce hospital costs [12, 13, 14]. Machine learning can predict the vulnerable population with social needs by using these EMR data and patient level data (claims/administrative data) after properly linking with the community-level geographical data. HRSNs prediction for social intervention design by machine learning is a novel approach. Holcomb et. al used Light GBM machine leaning algorithm to predict HRSNs of patients in the CMMI AHC Model from patient-level EMR data and publicly available community-level SDOH data [15]. The hyper parameters were tuned using Bayesian Optimization to get the best performing model. This study successfully detected patients with social needs in different domains and is summarized in Table 1.

**2.2. Disparity Index:**

Disparity, a major concern in today's healthcare, refers to the existing differences between racial and ethnic groups, gender, age, sexual orientation, disabilities, socio economic status. Healthcare disparities are differences in healthcare outcomes that are linked to social, economic, environmental disadvantage – often driven by the social conditions in which people live, learn, work and play. According to CDC, Health disparities are related to the historical and current unequal distribution of social, political, economic, and environmental resources.

Disparities in health and healthcare not only affect the groups facing disparities, but also limit overall improvements in quality of care and health for the broader population and result in unnecessary costs. As a result, people of color and other underserved groups experience higher rates of illness and death across a wide range of health conditions, limiting the overall health of the nation [16].That's why The Joint Commission (TJC) published new guidance and requirements in June 2022, shown in Table 1,  for reducing disparities in health care, taking effect in January 2023. Moreover, Healthy People 2030 provides tools for advancing health equity and helping individuals, organizations, and communities committed to improving health and well-being [17].

Machine learning has proven the most effective way to detect and manage disparities. Previous statistical analysis for detecting disparities has become obsolete due to the increasing dimensions of data and for data which is unknown because of sensitivity. Zahn et al., proposed a data driven framework Automatic location of Disparity (ALD) which can deal with both categorical and continuous predictors and handle disparities generated from complex and multi-way interactions of different groups [18]. Moreover, transfer learning can improve machine learning models to provide effective approach in reducing disparity among data-disadvantaged ethnic groups [19]. Even depression among different groups of adolescents: adolescents without any insurance, less authoritative parents, groups of negative school experiences are examined and predicted by multivariable logistic regression [20].

During Covid time, machine learning algorithms were also used in determining telemedicine and healthcare disparities. Multilevel logistic regression analysis is used in [21] to get the associations between each of the statistically significant variables and analyze patient-level and community-level covariates. ML also can be used in finding clinic level disparities which is very crucial for overall state performance, eventually national level performance. Even disparity in health care practitioner job scheduling also can be determined easily by the help of machine learning applications.

### 3. Risk Management / Preventive Care:

### *3.1. Hierarchical Condition Categories (HCC) Score:*

Value based care has created renewed interest towards the hierarchical condition categories (HCCs). HCCs are the risk adjustment model to identify patients with chronic conditions and are mainly used by CMS [22]. HCCs help to estimate payments to healthcare for patients who are insured by different affordable insurance plans. According to 2022 HCC coding and documentation, there are a total of 86 HCC codes and divided into nineteen categories [23]. Risk adjustment factors are calculated based on patients' disease burden and demographic information and associated with each HCC. The traditional large formula for risk adjustment calculation is replaced by machine learning algorithms and depends on the ordinary least squares method. But applying other machine learning algorithms like ridge regression also improves the score [24]. ML algorithms can also be used for variables selection in risk adjustment score calculation as fit performance of estimation is as good or better than the current one's [25]. All necessary information related to HCC screening is shown in Table 2.

**Table 1: Summarized Information on Healthcare's Community Level Works**

| Ref. | Document Type | Published Year | Category | Description |
|------|---------------|----------------|----------|-------------|
| [7] | Report | 2017 | AHC Screening | Details about AHC HRSN Screening Tool |
| [8] | Article | 2019 | AHC Screening | Clinicians' perspective towards social determinants of health screening |
| [9] | Issue Briefs & Report | 2011 | AHC Screening | Connection between social needs and good health |
| [10] | Article | 2018 | AHC Screening | Food insecurity relation with hospital visits |
| [11] | Article | 2021 | AHC Screening | Benefits of systematically addressing and assessing social needs |
| [12] | Article | 2020 | AHC Screening | Social determinants of health effect on diabetic patients |
| [13] | Article | 2019 | AHC Screening | Impact of early assessment and intervention on ED visits |
| [14] | Article | 2021 | AHC Screening | Housing needs impact in hospital utilization |
| [15] | Article | 2022 | AHC Screening | Predicting HRSNs by machine learning |
| [16] | Website | 2023 | Disparity Index | Recent status of disparity in health and healthcare with plausible solutions |
| [17] | Website | 2022 | Disparity Index | Healthy People 2030 goals: eliminate health disparities, achieve health equity, attain health literacy |
| [18] | Article | 2023 | Disparity Index | Automatic location of disparities (ALD) using ML |
| [18] | Article | 2020 | Disparity Index | Addressing data inequality and data distribution discrepancies among ethnic groups by transfer learning method of machine learning |
| [20] | Article | 2019 | Disparity Index | Risk factors and healthcare disparities of adolescent depression |
| [21] | Article | 2021 | Disparity Index | Racial and ethnic disparities in healthcare access via telemedicine during Covid-19 |

**Table 2: Summarized Information on Healthcare's Risk Management/Preventive Care**

| Ref. | Document Type | Published Year | Category | Description |
|------|---------------|----------------|----------|-------------|
| [22] | website | - | HCCs | HCCs and RAF's working principle |
| [23] | Report | 2022 | HCCs | 2022 HCC coding and documentation tips |
| [24] | Article | 2021 | HCCs & ML | Use of ML on HCCs and service-level propensity scores combination to reduce service-level selection |
| [25] | Article | 2021 | HCCs & ML | ML based formula with fewer variables to improve the performance of health plan payment systems |
| [26] | Website | 2021 | Mortality rate | Inpatient mortality indicators of California hospitals |
| [27] | Article | 2017 | Mortality rate | Identifying demographic pattern of mortality, time spent in ED before death and risk factors for fatal outcomes |
| [28] | Article | 2015 | Mortality rate | Mortality rate relation with quality of healthcare |

| [29] | Article | 2020 | Mortality rate & ML | Mortality prediction of patients by ML diagnosed with Covid-19 |
| [30] | Article | 2021 | Mortality rate & ML | Predicting mortality rate in patients with Covid-19 using ML |
| [31] | Article | 2020 | Mortality rate & ML | In-hospital mortality of sepsis patients in the ICU using ML |
| [32] | Article | 2020 | Mortality rate & ML | Mortality prediction of critically ill diabetic patient using ML and clinical notes |
| [33] | Article | 2019 | Mortality rate & ML | Mortality prediction after transcatheter Aortic valve replacement using ML |
| [34] | Article | 2019 | Mortality rate & ML | In-hospital mortality prediction of acute kidney injury ICU patients using random forest model |
| [35] | Article | 2018 | Mortality rate & ML | Burn patient's survival prediction using ML |
| [36] | Article | 2019 | Mortality rate & ML | Breast cancer patient's survival prediction using ML |
| [37] | Article | 2020 | Mortality rate & ML | In-hospital mortality prediction of patients with febrile neutropenia using ML |
| [38] | Article | 2021 | Mortality rate & ML | chemotherapy-induced ulcerative mucositis prediction using ML |
| [39] | Website | - | Surgical risk | National Surgical Quality Improvement Program |
| [40] | Article | 2018 | Surgical risk & ML | Surgical risk prediction of patients undergoing elective adult spinal deformity procedures using ML |
| [41] | Article | 2014 | Surgical risk & ML | postoperative pulmonary complications prediction |
| [42] | Article | 2008 | Surgical risk | Number of surgical procedure s in an American |
| [43] | Article | 2021 | Surgical risk & ML | Postoperative surgical complications prediction |
| [44] | Article | 2009 | Surgical risk | Pre-operative prediction of cardiac transplant cellular rejection by neutrophil response |
| [45] | Article | 2002 | Surgical risk | Pretransplant renal function impact on orthotopic liver transplantation |
| [46] | Article | 2015 | Surgical risk | Surgical complication prediction |
| [47] | Article | 2011 | Surgical risk | Bariatric surgery risk prediction |
| [48] | Article | 2015 | Surgical risk & ML | Pre-operative risk assessment of elderly ovarian cancer patients undergoing primary cytoreductive surgery |
| [49] | Article | 2018 | Surgical risk & ML | Complication prediction of Posterior Lumbar Spine Fusion using ML |
| [50] | Article | 2022 | Surgical risk & ML | Complication prediction of primary total knee arthroplasties using ML |
| [51] | Article | 2022 | Surgical risk & ML | Surgical complication prediction of abdominal wall reconstruction using ML |

## 3.2. Mortality Rate:

Mortality rate plays a vital role in health care industries. Inpatient mortality rate, developed by the federal Agency for Healthcare Research and Quality (AHRQ) indicates the efficiencies in the quality of hospital care [26]. But the mortality rate in the emergency department, which is the face of any hospital, evaluates the hospital performance [27]. Most of the hospitals, nowadays, use risk adjustment methods to identify any poor performance. Safety, effectiveness, emergency medicine, timeliness of care, effective resource management are reflected by the hospital mortality rate [28]. All admitted patients' demographic, characteristics, comorbidities, other factors are weighted to reflect the strength of association with mortality.

Machine learning algorithms are effective in predicting expected mortality rate which is then compared with the actual hospital mortality rate and helps to find the discrepancy between these two. Feature importance characteristics of ML algorithms can also determine the important factors responsible for such mortality rate and in turn, helps to take crucial management decisions. Optimal resource utilization, especially in intensive care units (ICUs) improves the clinical outcomes of a hospital. Especially during the Covid-19 pandemic, the shortage of hospital beds, critical care equipment, human resources emphasize the accurate prognosis prediction to triage patients effectively. Machine learning models based on pre-diagnosis information of infected individuals helped to allocate limited medial resources [29, 30, 31]. Different ML algorithms: least absolute shrinkage and selection operator (LASSO), random forest (RF), gradient boosting machine (GBM) and the traditional logistic regression (LR) can also be used in early and accurate identification of sepsis patients with high risk of in-hospital death [31].

Nowadays most of the diseases associated with high-risk complications, longer hospital stays, higher medical cost, and in-hospital death. Diabetics mellitus, which is one of the most common diseases characterized by chronic hyperglycemia, is increasing day by day and has led to an increase in the number of ICU patients. Machine learning and natural language processing (NLP) approaches are used to predict the mortality risk in chronic diabetes patients using clinical notes [32]. Transcatheter aortic valve replacement (TAVR) is considered the gold standard treatment for patients with severe symptomatic aortic stenosis. TAVR depends on ML Methods to generate robust models for predicting the in-hospital mortality of patients [33]. Early mortality prediction is especially important for hospitalized patients with acute kidney injury (AKI). These tools provide decision-making support for clinicians to avoid ineffective clinical intervention or non-optimal treatment. Random forest algorithm is used to predict mortality for acute kidney injury (AKI) patients in the ICU [34].

Burn patients, superficial burns to severe burns require complex care involving a delicate balance among resuscitation, stabilization, and rehabilitation. Survival of burn patients can be predicted using machine learning techniques in patient and hospital level [35]. Though patient factors are difficult to control, hospital factors help to decide the suitable hospitals for any burn patients as it is proven that hospital factors such as full-time residents (p < 0.001) and nurses (p = 0.004) to be associated with increased survival [35]. Moreover, breast cancer, most common cancer in women around the world needs machine learning techniques in detecting and visualizing significant prognostic indicators of survival rate [36]. Febrile neutropenia (FN) is a life-threatening complication of cytotoxic chemotherapy, associated with high mortality among cancer patients. The required intensive physicians' subjective evaluation in scoring indexes for FN mortality calculation can be replaced by machine learning models [37]. Additionally, Cancer treatment induced mucosal inflammation and the risk factors for mucosal inflammation are the spectrum of pathobiology, dose, drug, patient related. As Ulcerative mucositis (UM) has less recognized risk factors, most cancer therapists in these adverse events utilize machine learning approaches for predicting chemotherapy-induced UM [38].

### 3.3. Surgical Risk:

Measure and improve the quality of surgical care has become the crucial topic in healthcare industries in the US. More than six hundred hospitals have come together under the American College of Surgeons (ACS) National Surgical Quality Improvement Program (NSQIP). This data sharing collaboration makes it possible to discuss best practices and design quality improvement initiatives in surgery [39]. Utilization of this big data also helps in more precise risk stratify and prognosticate how an individual patient will behave given a disease or intervention [40]. Detecting significant associations between preoperative variables and postoperative outcomes and finding out the major clinical risk factors enhance informed consent before surgery, provide a guideline in clinical decision-making during the perioperative period, and measure the quality and safety of hospital care [41].

According to [42], shown in Table 2, an American encounters nine surgeries on an average in lifetime and each surgery has its own complications. Surgery complications are long term and difficult to compensate. Sometimes complications are so daring that cause death too. Moreover, identifying the surgical patients at risk of postoperative complications and providing personalized precision medicine-based treatment strategies is the only way to deal with the increasing health-care cost and limited financial resources [43]. The availability of data and applications of machine learning algorithms are creating a new era in post-surgery complication detection. Nowadays clinicians can predict the major risk factors associated with any surgery based on the readily available data and take preventive measures to avoid them.

Ability to determine the major outcomes of complex surgeries like neutrophil response of cardiac transplant [44], renal functioning after orthotopic liver transplantation [45], makes the machine learning and deep learning algorithms inseparable from healthcare research. In the general surgery department, the predictive factors used by machine learning can be categorized in four different groups: Patient-related factors, Co-morbidities, Laboratory values, and Surgery-related factors [46]. Variety of conditions grouped in Postoperative pulmonary complications (PPCs): pneumonia, aspiration pneumonitis, respiratory failure, reintubation within 48 h, weaning failure, pleural effusion, atelectasis, bronchospasm, and pneumothorax are the leading cause of death and hospital care expenditures in recent years [41]. Multivariate logistic analysis can be used to predict the PPCs successfully. The predictive scoring system of Bariatric surgery for severe obese patients can facilitate the selection of procedures and allow better risk stratification, especially in the case of high-risk patients [47].

Some surgeries need primary therapy to make the patient ready for surgery. Even this primary therapy can be tailored if the post complications of surgery can be detected earlier. The study [48] shows whether neoadjuvant chemotherapy followed by Cytoreductive surgery will have less complications than without therapy surgery in elderly ovarian cancer patients can be predicted prior by preoperatively identified unresectable disease, significant medical co-morbidities, or poor performance status. Complications, irregular surgery duration, and major risk factors following adult spinal deformity [40], lumber spine fusion [49], total knee arthroplasties [50] can also be predicted using machine learning and artificial neural network algorithms. Patient specific risk assessment for finding the factors behind adverse outcome of any surgery like AWR (abdominal wall reconstruction) and hernia recurrences [51] is only possible today for the advancement of machine learning algorithms.

## 4. Healthcare Operation Management:

### 4.1. Resource management:

Proper utilization of healthcare resources ensures the quality of care and in-time delivery of services. Appropriate usage of available medical staff and professionals is extremely hard to overstate [52]. Compressed work arrangements are extremely popular but not suitable for healthcare because in healthcare many constraints like healthcare history and culture, empathy, ethical reflection have to be considered. The requirement of negotiating a large dimension of constraints to ensure flexible working conditions can only be possible by the application of machine learning which can mine user-defined and soft constraints and transform staff scheduling a classification problem with as high as 93.1% accuracy [53].

Another major application of machine learning in healthcare is a conversational agent named chatbots which holds conversations with humans, represent a healthcare practitioner, in-real time for providing timely support, human-like knowledge transfer and communication. Complex dialog management and conversation flexibility make chatbots more acceptable to beneficiaries in recent times. Nowadays chatbots are trained to become specialized in specific diseases like oncology, diabetics and involve in-depth discussions and examples of diagnosis, treatment, monitoring, patient support, workflow efficiency, and health promotion [54]. From making an appointment with a doctor to disease prediction and treatment recommendation, chatbots are being widely used. Especially, it became more popular during Covid-19 pandemic due to the high-volume support management it provided. Moreover, machine learning has already been using in crucial resources like hospital bed, ICU, medication, vaccination, medical supplies management and well-known for patients and hospitals cost reduction. Beforehand prediction of vulnerable patients not only utilize specialized doctors and nurses but also adjust the risk scores in precision payment reception from insurance companies which is significant in maintaining overall ongoing operations of healthcare. Different types of work on healthcare resource management are shown in Table 3.

**Table 3: Summarized Information on Healthcare Operation Management**

| Ref | Document Type | Published Year | Category | Description |
|---|---|---|---|---|
| [52] | Website | - | Resource utilization | Importance of proper resource utilization in nursing |
| [53] | Article | 2021 | Resource utilization | Employee scheduling using ML classification algorithm |
| [54] | Article | 2021 | Resource utilization | Chatbot for healthcare focusing on oncology using ML |
| [55] | Article | 2023 | Patient classification & ML | Patient classification using basic vital signs in emergency department by ML |
| [56] | Article | 2020 | Patient classification & ML | Patient subgrouping with latent disease discovery using electronic health record by ML |
| [57] | Article | 2019 | Patient classification & ML | Usage of patient clustering to predict mortality and hospital stay using electronic health record by ML |
| [58] | Article | 2021 | Patient activity monitoring & ML | Covid-19 patient monitoring using ML |
| [59] | Article | 2019 | Patient activity monitoring & ML | Autonomous ICU patient monitoring using pervasive sensing and deep learning |
| [60] | Article | 2022 | Patient activity monitoring & ML | Fall detection in healthcare using smartwatch sensors and deep learning |
| [61] | Article | 2019 | Patient activity monitoring & ML | Diagnosis and monitoring Alzheimer's patients using body worn inertial sensors and ML |
| [62] | Article | 2020 | Patient activity monitoring & ML | Diabetic patient monitoring using smart devices and sensors by ML |
| [63] | Article | 2019 | Patient activity monitoring & ML | Use of activity tracker data by ML to classify cohort of heart patients |
| [64] | Article | 2017 | Patient activity monitoring & ML | Self-harming activity monitoring by in-patients in clinical settings |
| [65] | Article | 2022 | Patient activity monitoring & ML | Face-touching activity monitoring using wristwatch sensors by ML |
| [66] | Website | 2018 | Patient Experience Analysis | Common elements of patient-centered care plans |
| [67] | Article | 2014 | Patient Experience Analysis | Patient experience importance in high-quality medical care and its relationship to other domain of quality |
| [68] | Article | 2003 | Patient Experience Analysis | Challenges and differences of public reporting with recommendations for effectiveness in US and UK |
| [69] | Article | 2016 | Patient Experience Analysis | Patient experience impacts on quality improvement of healthcare |
| [70] | Article | 2013 | Patient Experience Analysis | Online comment analysis for capturing patient experience |
| [71] | Article | 2022 | Patient Experience Analysis & ML | Patient experience understanding analyzing comments on transitions of care using ML |
| [72] | Article | 2020 | Patient Experience Analysis & ML | Drug review classification using deep learning and ML |
| [73] | Article | 2021 | Patient Experience Analysis & ML | Patient satisfaction analysis using satisfaction survey and EMR data by interpretable ML |
| [74] | Article | 2019 | Patient Experience Analysis & ML | Dry eye disease patient experience evaluating using NLP |
| [75] | Article | 2020 | Patient Experience Analysis & ML | Dermatology patient experience evaluating using reddit data and NLP |
| [76] | Article | 2009 | Patient Experience Analysis | Staff and inpatient feedback impact on patient experience improvement |

### 4.2. Patient categorization/ subgrouping:

Patients of different conditions encounter in the emergency department and make it difficult to manage all of them together in rush hours. In this case, patient classification by ML according to their critical conditions determines if the patient needs immediate medical intervention by clinicians [55] or transfer to inpatient or outpatient department. Clustering also identifies patients' subgroups based on similarity and ensures appropriate distribution of patients to doctors based on specialization. That is why unsupervised machine learning algorithms are highly effective in-patient categorization and latent disease clusters discovered [56]. Even length of stay of patients can be determined more accurately than before by clustering EMR data into several communities and simultaneously training one ML model per community [57].

### 4.3. Patient Activity Monitoring:

Regular monitoring of the patients in the healthcare is incredibly significant but difficult by the overburden nurses. Heath status of the patient, especially the critical ones, can be changed in no time. But it is not possible to monitor all the patients at granular level even if there are enough personnel. The introduction of pervasive computing in combination with machine learning brings miracles in the field of patient activity monitoring. In the near future, nurses will only be notified when the patient's status is about to change and need human intervention. The advancement of internet of things (IoT) and different types of sensors are creating piles of data and making opportunities of abnormal pattern extraction by machine learning models. The abundance of devices like smartwatch, mobiles, cameras, sound sensors, light sensors and their pervasiveness are making the ubiquitous monitoring system for patients. The importance of remote activity monitoring systems is nowadays well-known because of their widely use during Covid-9 pandemic to curb the spread of the virus and to check the health status of Covid affected patients [58].

Moreover, Davoudi et. al demonstrated the autonomous monitoring system, referenced in Table 3, for critical ICU patients using different types of available devices and proved that granular level monitoring is possible using noninvasive system [59]. Falls represent one of the major risk factors in the case of elderly patients in healthcare. Nowadays different types of wearable devices having accelerometer and gyroscope sensors can detect falls leveraging machine learning algorithms [60]. Daily activity monitoring of patients having Alzheimer disease creates activity profile to evaluate the vulnerabilities of patients based on activity level which in turn shows up to 82% improvement in compared to other existing techniques [61]. Health status of diabetic patients [62], patients with heart diseases [63] are autonomously monitored by fitness trackers and report the changes to responsible person even if that person is not nearby. Additionally, this activity monitoring system particularly helps self-harming tendencies detection in mental disorder patients [64]. It also helps the patients who had eye surgery recently by alerting them in case of hand transition towards eye [65].

### 4.4. Patients Experience Analysis:

Today's healthcare is inclined to patient centric services; more responsive and respectful of patient's experiences, values, needs and preferences than before [66]. Providing high-quality medical care depends on the assurance that patients value guided all medical decisions which ultimately depends on their experience. According to Dr. Foster Intelligence, patient experience is the "feedback from patients on 'what actually happened' in the course of receiving care or treatment, both the objective facts and their subjective views of it." Policy makers around the world are now emphasizing gathering patient experience data to assess the providers against a range of performance indicators to stimulate quality improvement [67]. Different studies found a positive association between clinical effectiveness and patient safety with patient experiences.

Patient experience depends on multiple dimensions of care starting from the process of making or receiving an appointment, cleanliness of facilities, waiting times, the information provided to interactions with staff including receptionists, healthcare assistants, nurses, and doctors [67]. It is also evident that public reporting of patient experience data with the details of intervention, taken into consideration in the context of healthcare system can be amazingly effective in stimulating providers to improve their quality of care [68]. So, more attention should be given to how the patient's experience reflects changes in practice and in turn, these changes may have an impact on patient experience in future [69].

Although patient experiences are now in the center of healthcare services, it is exceedingly difficult to understand the complex nature of the data because same experience may sometimes interpret differently based on the expectations of different population groups [67]. Moreover, though data are available in all stages of care, even on the internet, it is cumbersome to parse this large amount of unstructured, free text information for gaining any meaningful insights. But due to the use of natural language processing (NLP) and machine learning, the theory of patient experiences impact on healthcare are being possible to incorporate in real life application. Categorizing patients' comments in positive and negative descriptions NLP, ML automatically predicts whether the patient recommended a hospital, whether the hospital was clean and treated with dignity [70].

Patient sentiment analysis by NLP and ML models based on patient experience survey comments relating to transitions of care in all four healthcare settings: accident and emergency (A&E), inpatient, outpatient and maternity, highlights some of the problems in care transitions. Support Vector Machine (SVM) identifies "discharge" in inpatients and Accident and Emergency, "appointment" in outpatients, and "home' in maternity have negative sentiments in transition and continuity [71]. In case of drug assessment on patients, use of ML in drug reviews analysis makes it possible to understand the patient feelings towards drugs usage more easily than before by practitioners [72]. Interpretable machine learning helps in building trust and confidence in health service providers and formulating patient satisfaction problems altogether. It can identify the most influential factors in patient satisfaction and transform heterogeneous data into human understandable features [73].

Use of machine learning in online platforms for patient experience and sentiment analysis is extremely popular nowadays. Social media listening (SML), looking at what people discuss in online social media forums about their disease, in association with ML algorithms can be utilized to inform early drug development process, market access strategies and stakeholder discussions [74]. One study [75], shows the potential of reddit data source in gaining insights of patients interested in dermatology. NLP usage on this forum not only helps doctors to learn about the therapeutic options that patients might have tried but are reluctant to bring up in clinic visit but also learn about the treatments under discussion in the community. Though there is plenty of patient experience data in hospital facilities and internet, according to [76], we can also use staff feedback data because staff feedback is associated with patient-reported experience. In case the patient data needs complex preprocessing for analysis, management, clinicians, and other staff should feed the staff survey in ML models to improve the quality, safety, and patient experience. Table 3 shows all relevant references related to patient experience analysis.

## 5. Remote Care:

### 5.1. Telemedicine:

Telehealth, known as telemedicine, is the way to communicate with healthcare practitioners through the internet using mobile, computer, table etc. It is becoming popular day by day, especially during Covid-19 pandemic when physical contact was restricted, for its advantages over traditional doctor patient meet up system. Patients can meet doctors in their comfortable home environment. At the same time, doctors are able to do better assessment because they can see the patients in their home environment and explore the surroundings to find out the disease originating source [77]. It is also good to have family members around during check-ups who can provide more information about lifestyle. This type of virtual visit makes easy the regular check-up and helps maintain chronic conditions. But still there are some drawbacks. It is difficult to assess a patient who is not accustomed to modern technologies. Machine learning based computer-aided decision support system assists clinicians in disease detection, tracking disease progression, even suitable resuscitation prior to the arrival of patients to the hospital [78]. The utility of telemedicine in filtering serious patients to make in person appointments during pandemic times is irrefutable.

Moreover, correct medical treatment needs lots of medical data text, images, audio signals to be transferred between doctor and patient during appointment hours through the internet. Biomedical signals are sensitive in nature as a little noise can change the outcome of the test results and machine learning models are used to regenerate the signals under transmission more precisely [79]. Additionally, it becomes difficult for patients to address appropriate wording for their problem list. ML algorithms can perform a tremendous job here by suggesting related problems analyzing patients' conversations. It is even true for clinicians when they are struggling to come up with layperson terms to make patients understand more clearly. ML also assists healthcare practitioners to make high-quality decisions by finding the determinants of patients' satisfaction with telemedicine [80]. Table 4 summarizes the pertinent references on telemedicine.

**Table 4: Summarized Information on Healthcare's Remote Care**

| Ref | Document Type | Published Year | Category | Description |
|-----|---------------|----------------|----------|-------------|
| [77] | Website | - | Telemedicine | Benefits of telemedicine |
| [78] | Article | 2021 | Telemedicine & ML | Use of ML in electronic emergency triage and patient priority system in telemedicine |
| [79] | Article | 2019 | Telemedicine & ML | Biomedical signal transmission and regeneration in telemedicine using ML |
| [80] | Article | 2021 | Telemedicine & ML | Prediction of patient satisfaction with telemedicine using ML |

## 6. Early Detection:

### 6.1. Mental Health Detection:

Behavioral health refers to the behavioral impact on health and mental well-being. It studies the variables between our daily habits and their effect on physical health and also includes mental health disorders [81]. The prevention and treatment of mental health have become the major concern nowadays as according to the 2020 National Survey on Drug Use and Health, mental health conditions such as depression, anxiety, or schizophrenia prevails in 21 percent US adults and 17 percent of total youth has major depressive episode [82]. Due to its severity, behavioral health is also integrated with other health care and becoming especially important for maintaining proper life balance.

But poor access to behavioral health services is a hindrance in the continual behavioral monitoring and treatment where Machine learning is doing a tremendous job. Nowadays machine learning is used as a monitoring tool for individuals with deviant behavior [83]. Mental health status detection using designed questionnaire leveraged the clustering unsupervised algorithm for data labeling and different machine learning algorithms: support vector machines, decision trees, naïve bayes classifier, K-nearest neighbor classifier and logistic regression for identifying state of mental health [83].

In assessing patient satisfaction receiving mental health services (MHS), Andersen's Behavioral Model integrates sociodemographic, clinical, needs-related, service utilization, social support, and quality-of-life (QOL) variables and machine learning algorithms are used to explore different complex association between the dependent and independent variables [84]. Sometimes experts are unable to keep patients in constant observation and find quantification of behavior quite challenging. Behavioral signal processing with machine learning can augment their ability by quantifying behavior objectively and monitoring the behavioral construct over time [85]. Ability to measure person's internal state and high availability, low-cost physiological wearables are becoming increasingly popular. The medical Internet of Things (mIoT) is projected to be worth USD 94.2 billion by 2026 [86].

Recent studies also develop predictive models on mobile phone sensors data to monitor individual's mental wellbeing and notify concerned persons [87]. It holds great promise in clinical health research and very fruitful in monitoring at-risk populations. Mobile health interventions are becoming promising as most of the mental health patients need continuous observation by experts [88] and helping the clinicians to take decisions by providing continuum lifestyle data. Moreover, the use of deep learning in healthcare creates lots of hope in mental health prediction. Classification of neuroimages, electroencephalogram data, electronic health records, genetic data, Vocal and visual expression data helping healthcare professionals in making the right choice for diagnosis and prognosis [89]. All related information is referenced in Table 5 for quick look-up.

### 6.2. Early Disease Prediction:

Early detection of disease helps to take proper treatment plan before going into an advanced stage. Pattern recognition capabilities of ML algorithms from different types of text, image, audio, video data created the opportunity to predict a disease in incredibly early stage after comparing with labeled historical data. Skin cancer, more curable in the beginning stage, tends to spread gradually over other body parts. But analyzing the lesion parameters of affected area using deep learning algorithms, shown in Table 5, makes it possible to distinguish melanoma and benign skin cancer [90] and early detection. Early detection of scalp diseases [91] and other skin diseases are also possible for convolutional neural network. Diabetic kidney disease early prediction by monitoring of potential biomarkers like galectin-3 and GDF-15 [92]. Moreover, early detection of Parkinson's disease [93], Rheumatoid Arthritis [94], Diabetic Retinopathy [95] helps doctors to take early decision for advanced clinical tests and prevents the extreme cases from happening.

**Table 5: Summarized Information on Early Detection**

| References | Document Type | Published Year | Category | Description |
|---|---|---|---|---|
| [81] | Website | 2018 | Mental health detection | Difference between behavioral health and mental health |
| [82] | Website | 2022 | Mental health detection | Discussion on behavioral healthcare conditions in US |
| [83] | Article | 2018 | Mental health detection & ML | State of mental health detection by behavioral modeling using ML |
| [84] | Article | 2017 | Mental health detection | Assessing patient satisfaction with mental health services and identifying associated variables |
| [85] | Article | 2017 | Mental health detection & ML | Mental state detection using body sensors data by ML |
| [87] | Article | 2022 | Mental health detection & ML | Assessment of ML model generalizability in mental health symptom detection using mobile sensing data |
| [88] | Article | 2017 | Mental health detection & ML | Review on mental health research using personal sensing data |
| [89] | Article | 2020 | Mental health detection & ML | Review on DL applications in mental health research |
| [90] | Article | 2021 | Disease early detection & ML | Systematic review on skin cancer detection using deep learning |
| [91] | Article | 2023 | Disease early detection & ML | ML based hair and scalp disease detection |
| [92] | Article | 2020 | Disease early detection | Exploration of potential biomarkers for early diabetic kidney disease detection |
| [93] | Article | 2022 | Disease early detection & ML | Computing risk of bias automatically in early Parkinson's disease detection by artificial intelligence |
| [94] | Article | 2020 | Disease early detection | Electrochemical Immunosensor for the Early Detection of Rheumatoid Arthritis Biomarker |
| [95] | Article | 2020 | Disease early detection & ML | Early detection of diabetic retinopathy using deep learning |
| [96] | Article | 2018 | Food diet suggestion | Challenges behind the low food intake in hospital |
| [97] | Article | 2020 | Food diet suggestion & ML | ML based diet recommendation to patients |
| [98] | Article | 2021 | Food diet suggestion & ML | A systematic review on ML applications in food intake assessment |
| [99] | Article | 2022 | Food diet suggestion & ML | Resource on ML in nutrition research |
| [100] | Article | 2020 | Food diet suggestion & ML | ML based patient diet recommender |
| [101] | Article | 2020 | Food diet suggestion & ML | Personalized diabetes blood glucose control application using reinforcement learning |
| [102] | Website | - | Food diet suggestion | Discussion on precision nutrition |
| [104] | Article | 2020 | Food diet suggestion | Preoperative & postoperative diet plan impact on enhanced recovery after surgery among Surgical Gynecologic Cancer Patients |

### 6.3. Food Diet Planning:

Food intake has a great impact on body and mind for both healthy and unhealthy people. Specially in-hospital patients, robust diet recommended by dietician can significantly increase longevity, protect against further disease, improve quality of life, and reduce negative outcomes like readmission, prolonged hospital stay, mortality and hospital cost [96, 97]. Nowadays food intake evaluation has become a crucial part of scientific research and clinical trials as the same diet may have different effects on different individuals or population groups based on the variation in demographic and health conditions [98]. The complex relationship between nutrition and disease is difficult to capture fully because of so many influencing factors [99].

But due to the availability of tons of complex food data and machine learning algorithms, data-driven models are now able to fully reflect the overall scenario considering all other variables. Current screening tools in hospitals applying naive machine learning algorithms like K-means to classify the inpatients by their nutrition status [99]. This decision system is also helping healthcare in staff management, automating approaches. Machine learning algorithms correlate patient socioeconomic, demographic, health characteristics, institutional factors with their food intakes [96]. Many ML algorithms e.g., Logistic regression, decision tree, random forest, recurrent neural network, gated recurrent unit, long short-term memory is being used in dieting recommendation system. Deep learning classifiers are improving preciseness and accuracy by incorporating k-clique [100]. Reinforcement learning (RL), an automated goal-directed learning, is also being used in personalized dosage control based exclusively on patients own data [101].

Precision nutrition, a new trend in nutrition research, is the individual DNA, microbiome, and metabolic response to specific dietary plans are becoming popular for electing effective eating plans to prevent or treat diseases [102]. This can be a solution to many nutrition-related non-communicable diseases, such as obesity, diabetes, cancer, chronic vascular disease referenced in Table 5. ML in precision nutrition gives patient-level monitoring and metabolomic signature differences to identify the causal relationship between food intake and individuals' different metabolic responses. Moreover, dietary assessment to define the true intake rather than the estimated intake using image, video, sensors data of wristwatch or mobile phones is another field of ML in nutrition research [99]. Both patients and health care professionals are benefitting from patient self-management applications which deploy complex ML algorithms. In case of diabetic management, burden-free continuous remote monitoring for patient's symptoms and biomarkers produce better glycemic control with reductions in fasting and postprandial glucose levels, glucose excursions, and glycosylated hemoglobin [103].

Most of the time patients need to follow a specific diet plan before surgery or therapy. This helps to avoid complications and enhance recovery after surgery (ERAS). ERAS protocol applications significantly shorten the length of hospital stays (LHOS) and readmission in case of colorectal surgeries, gastrointestinal surgery, and various other fields [104]. The conventional feeding strategy of prolonged fasting might be a case of excessive catabolism result in weight loss and muscle mass loss. ERAS can leverage ML algorithms for producing a couple of weeks diet plan after surgery like radical cystectomy with or without neobladder based on the patient responses [105]. Generating patient satisfaction by showing the reasons for selecting specific foods are also possible due to explainable AI (EXAI) incorporating SHAP, LIME, and different ML algorithms.

## CONCLUSION

In this paper we have explored the existing ML applications in various areas of health sectors. Instead of technical jargons, we used some Layman terms and provided some easily understandable examples to make this work more relevant to a wider audience. While the paper mainly focuses on the Ongoing practices of ML in healthcare, it also sheds some light on the ML Research Development in healthcare which is not implemented widely yet but has huge potential in the near future. Thus, undertaking future development and improvement in the current healthcare system needs more data availability and ML models effective implementation. Lastly, there is a need for well-arranged bibliography on the cutting edge and existing work of ML in the healthcare system to encourage healthcare professionals in making modern healthcare systems.